\documentstyle[psfig,aclap]{article}
\title{Some apparently disjoint aims and requirements for grammar 
development environments:
the case of natural language generation}   
\author{John Bateman \\ Language and Communication Research\\
University of Stirling, Stirling, UK\\ ({\tt j.a.bateman@stir.ac.uk})}
\begin{document}
\bibliographystyle{acl}
\maketitle

\vspace*{-3in}{\fbox{\parbox[t]{4in}{\small From: ACL-EACL'97 Workshop
on {\em Computational Environments for Grammar Development and Linguistic
Engineering} (eds: Estival, Lavelli, Netter and Pianesi), Madrid, 1997.
pp1--8.}}}

\vspace*{+2.5in}

\begin{abstract}
Grammar  development    environments (GDE's)   for  analysis   and for
generation   have  not yet come   together.    Despite  the fact  that
analysis-oriented GDE's (such as {\sc alep}) may include some possibility of
sentence generation, the development techniques and kinds of resources
suggested are apparently not those required for practical, large-scale
natural language generation work.   Indeed, there is  {\em no  use} of
`standard'      (i.e.,    analysis-oriented)    GDE's   in     current
projects/applications targetting  the  generation of  fluent, coherent
texts.   This   unsatisfactory situation  requires  some  analysis and
explanation, which this paper attempts using as an example an
extensive GDE for generation. The support provided for distributed large-scale
grammar development, multilinguality, and resource maintenance are
discussed and contrasted with analysis-oriented approaches.
\end{abstract}

\section{Introduction: a problem}

Grammar  development  environments  (GDE's)   for  analysis  and   for
generation  have not yet come   together.   In fact,  the mainstay  of
design for linguistic resource development environments is skewed very
far from  that necessary for generation; this   is illustrated well by
the  following remark from an {\sc  eagles} (Expert Advisory Group for
Language  Engineering  Standards)    report   describing the   current
``convergence'' of opinion concerning  the required  functionality for
development platforms:
\begin{quote} ``The necessary
  functionality of a development platform  is more or less agreed upon
  by grammar writers.   \ldots They should  have a parser  for testing
  the developed grammars with respect to an input string, and possibly
  also           a               generator     to        test      for
  overgeneration.''~\cite[p117]{Eagles-fwg96}     
\end{quote}     
This  marginalization of the   generation process naturally impacts on
the kinds of  development and debugging  tools that are provided.  For
example,  perhaps the most  extensive  workbench developed within  the
European  Union, the Advanced Language Engineering Platform ({\sc alep}: cf.~\cite{Simpkins-etal93}),  while  forced   to   adopt  a
so-called `lean' formalism in  order to achieve acceptable efficiency,
nevertheless orients  itself most  closely to `mainstream'  linguistic
formalisms such  as HPSG and   LFG.  Neither of  these formalisms have
however found  widespread   use in larger-scale generation   contexts. 

There also continue  to be substantial  projects  whose specific goals
are to    build    or collect   linguistic  resources    for  language
engineering---including,  for   example,  projects  such  as Acquilex,
Eagles,              TransTerm,             EuroWordNet            and
others.\footnote{Sch\"utz~\cite{Schuetz96-amta}   provides   a    useful
overview of      current   language     engineering    projects  where
multilinguality  plays a role.}  However,   these projects have  not
apparently been  configured  to provide the  kinds  of  resources that
generation requires.  This can be  seen in the virtually zero  take-up
of such `mainstream' (i.e., analysis-oriented) resources in generation
projects  (both monolingual and multilingual)  where the goal has been
to provide efficient generation of realistic, useful texts.


Thus,  not only  is there a   lack of uptake of  linguistic resources,
there    is also   virtually    {\em  no   use} of  `standard'  (i.e.,
analysis-oriented)  GDE's in  current projects/applications targetting
the  generation  of   fluent,  coherent texts.    This  unsatisfactory
situation  certainly requires  some  analysis and  explanation---which
this paper  attempts.   To do  this, we  first briefly illustrate  our
claim that the grammar development environments  and approaches
that are  adopted  in natural   language generation  are by  and large
disjoint to those developed in natural language analysis. We then show
how  the  main  property  that  effectively  distinguishes  successful
generation grammars  from analysis  grammars  (regardless of  what the
grammars are used for) is  their orientation to communicative-function
and that it is precisely this property that plays a crucial role in creating
powerful and efficient grammar development environments that {\em are}
suited to the generation  task. 

We  illustrate  this relationship   between  resource organization and
development   tools by  focusing  on  techniques   for  developing and
maintaining    large-scale linguistic  resources   (mostly grammar and
semantics-grammar  mappings), for distributed grammar development, and
for supporting  multilinguality  that  have developed  for  generation
work.  A  direct question raised by  the paper is  then the  extent to
which the techniques  discussed could also  be relevant and applicable
to analysis-oriented development environments.

\section{The lack of use of analysis-based GDE's for generation}

There is clearly a partially `sociological' explanation to the lack of
exchange between approaches   in analysis and generation: the   groups
working on analysis and text generation are by  and large disjoint and
the questions and issues thus central in these groups are also at best
only partially overlapping.  This  is not, however,  sufficient.  Most
input   to analysis-oriented  work (e.g.,~\cite{ET-6-finalreport}) has
attempted  to   achieve a workable  level    of generality and  formal
well-foundedness that would guarantee the widespread applicability and
re-usability of  their results.  If this were  sufficient and had been
successful,  one  could expect generation   developers to have availed
themselves of  these results.  But  uptake for generation continues to
be  restricted to  those working  in the analysis-oriented  tradition,
mostly in the pursuit  of `bi-directional' sentence generation on  the
basis  of resources developed   primarily  for analysis.  `Core'  text
generation activities remain untouched.

\begin{table*} 
\rule{\textwidth}{0.2mm}
\begin{center}\small
\begin{tabular}{|r||c|c|c||c||c||} \cline{2-6}
\multicolumn{1}{r||}{\ } & 
\multicolumn{2}{c|}{\bf functional approaches} &
\parbox{1in}{\bf dependency approach} &
\parbox{1in}{\bf template approach} &
\parbox{1in}{\bf structural approach} \\ 
\multicolumn{1}{r||}{\ } & KPML/Penman & FUF & MTM & & \\ \hline \hline
\parbox[t]{0.6in}{\bf connected texts (differing text types)} & 
\parbox[t]{0.6in}{\em TechDoc\\ \rm HealthDoc\\ \em KOMET \\ $\vdots$} & 
-- & -- & Peba-II & -- \\ \hline
\parbox[t]{0.6in}{\bf connected texts (single text type)} &
\parbox[t]{0.6in}{\em GIST\\ Drafter\\ AGILE\\ $\vdots$} & 
\parbox[t]{0.6in}{PlanDoc\\Streak\\Comet\\ \em GIST\\ $\vdots$} &
\parbox[t]{0.6in}{\em FOG\\ LFS\\ MultiMeteo} &
\parbox[t]{0.6in}{(ILEX)} &
\parbox[t]{0.6in}{WIP\\({\em GIST})}  \\ \hline
\parbox[t]{0.6in}{\bf single sentences / utterances} &
\parbox[t]{0.6in}{Speak!\\ \em Pangloss} &
\parbox[t]{0.6in}{some} &
-- &
\parbox[t]{0.6in}{several} &
\parbox[t]{0.6in}{\em Verbmobil\\ CLE \\ \rm IDAS} \\ \hline
\parbox[t]{0.6in}{\bf single phrases} &
-- & some & many  & many & \parbox[t]{0.6in}{\em ANTHEM} \\ 
\hline \hline 
\end{tabular}
\end{center}
\begin{quote}Projects given in italics are essentially
multilingual---i.e., they are concerned with the generation of texts
in at least two languages.
\end{quote}

\caption{Distribution of generation systems by task and approach}
\label{gen-overview}
\rule{\textwidth}{0.2mm}
\end{table*}

One,   more  contentful,  reason  for  this    is that  the particular
requirements of   generation  favour an  organization  of   linguistic
resources  that has itself proved  supportive  of powerful development
and  generation    environments.   To   clarify the needs  of generation  and the relation   to the GDE's
adopted,  we  can  cross-classify approaches  adopted   for generation
according    to the  kind  of    generation  targetted.  This  largely
corresponds to  the {\em size} of  linguistic unit generated.  Thus we
can usefully distinguish  generation of single phrases,  generation of
single sentence or   utterance generation (such  as  might also  still
occur in  MT most  typically  or in utterance  generation in  dialogue
systems), generation  of connected  texts  of a single  selected  text
type, and generation of connected  texts of  a  variety of text  types
(e.g., showing variation for  levels  of user expertise, etc.).  These
are  distinguished precisely because it  is well known from generation
work  that  different   issues  play a   role  for   these   differing
functionalities.

Three   generation  `environments' cover   the   majority of  projects
concerned with text   generation where generation for  some  practical
purpose(s) is  the  main aim, not  the  development of some particular
linguistic  treatment or pure research  into problems of generation or
NLP   generally.  These   are   Elhadad's~\cite{Elhadad90-types}      `Functional
Unification   Formalism'     ({\sc fuf}),    the       {\sc     kpml}/Penman
systems~\cite{MannMatthiessen85-benson,KPML1},  and  approaches within
the Meaning-Text Model (cf.~\cite{MelcukZholkovskij70}) as used in the
CoGenTex-family of generators. Here resources   appropriate  for  real  generation are
accordingly understood as broad  coverage (with  respect to a
target application or set of  applications) linguistic descriptions of
languages that provide mappings  from enriched semantic specifications
(including details of  communicative effects and textual organization)
to corresponding  surface strings  in close  to  real-time.  
In addition, there  are many systems
that adopt in contrast  a template-based approach to  generation---now
often combined with full generation in  so-called `hybrid' frameworks. 
While,  finally,  there is  a very small  number  of serious, large-scale
and/or practical projects where analysis-derived grammatical resources
are     adopted.      This      distribution   is    summarized     in
Table~\ref{gen-overview}.   Importantly,  it   is {\em  only}  for the
approaches in the final  righthand column that standard analysis-based
GDE's  appear   to   be  preferred or   even  applicable.\footnote{For
  references to  individual systems see  the Web or a detailed current
  state of the art such  as Zock and  Adorni~\cite{ZockAdorni96-intro}
  or Bateman~\cite{Bateman-ADG}.}

\section{Communicative function: a common thread in generation resources}

It is well known in natural  language generation (NLG) that functional
information concerning  the communicative   intent of some   utterance
provides a   convenient   and  necessary organization  for   generator
decisions (cf.~\cite{McDonald80,Appelt85,McKeown85-book}).
Different approaches focus  the  role of  communcative functions  to a
greater  or less degree.   Some subordinate  it entirely to structure,
some  attempt to  combine structure and  function felicitously, others
place communicative function  clearly in the foreground.  Among  these
latter, approaches based on systemic-functional linguistics (SFL) have
found   the    widest application.  Both     the   {\sc fuf} and {\sc kpml}/Penman
environments draw heavily on SFL. This is to emphasize the role of the
{\em  paradigmatic} organization   of resources in  contrast to  their
syntagmatic, structural  organization. It turns out that it  is this crucial distinction
that provides  the cleanest  account  of the  difference between a GDE
such as {\sc alep} and one such as {\sc kpml}.

Viewed formally, a paradigmatic description of grammar such as that of
SFL attempts  to place as much of  the work of  the description in the
type lattice constituting  the grammar.  The  role of constraints over
possible feature structures   is minimal.  Moreover,  the distinctions
represented in the  type  lattice  subsume  all kinds  of  grammatical
variation---including variations that in,  for example,  an HPSG-style
account might be considered as examples  of the application of lexical
rules.  Diathesis alternations     are one clear    example; differing
focusing constructions are another. These are all folded into the type
lattice. Generation with such a resource is then reduced to traversing
the    type lattice, generally   from  least-specific to most-specific
types, collecting  constraints  on structure.  A   grammatical unit is
then exhaustively  described by  the complete  list of types  selected
during   a traversal:  this is  called  a  {\em selection expression}.
Additional mechanisms (in particular, the `choosers') serve to enforce
determinism: that is, rather than  collect parallel compatible partial
selection   expressions,   deterministic generation  is   enforced  by
appealing  to  semantic   or  lexical    information   as and     when
required.  This approach, which is theoretically  less  than ideal, in
fact supports  quite efficient generation. It  can be equated with the
use of `lean' formalisms in analysis-oriented GDE's.

\begin{figure*}
\rule{\textwidth}{0.2mm}
\centerline{\hbox{\psfig{figure=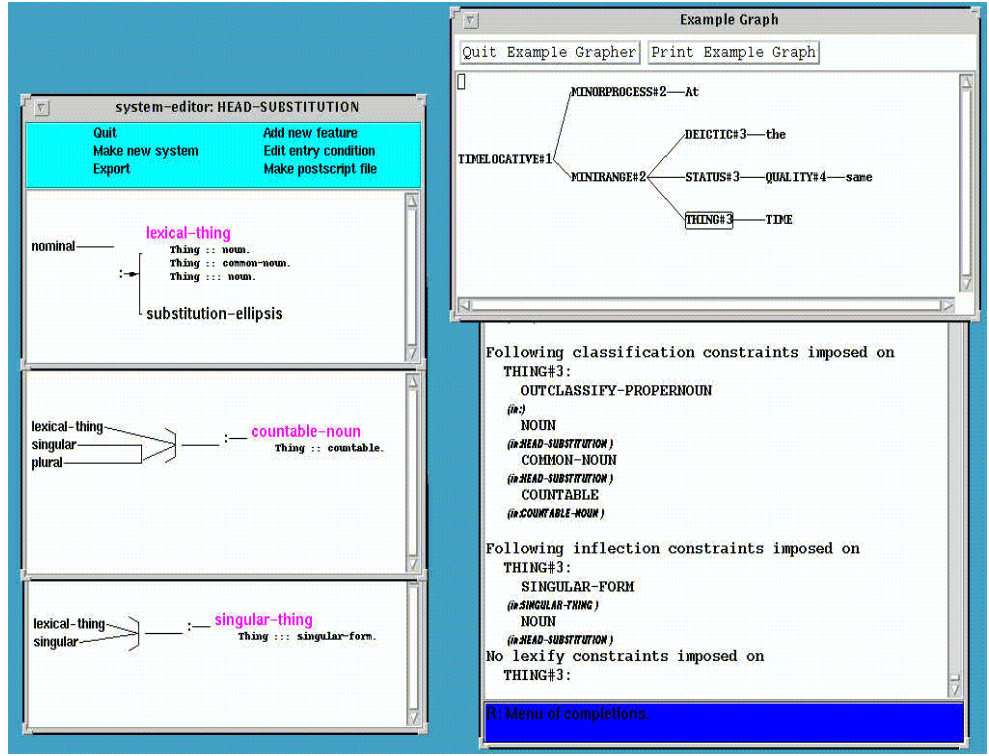,height=100mm,width=130mm}}}
\caption{Accessing points in the grammatical type lattice from
a generated structure}
\label{constraints-to-systems}
\rule{\textwidth}{0.2mm}
\end{figure*}

This paradigmatic design sketched here has  proved to have significant
consequences     for   the     design   of     appropriate development
environments. The   properties of these development   environments are
also  directly  inferable  from   the  properties of  the  linguistic
descriptions they are to support.  Among the results are:
\begin{itemize}\setlength{\itemsep}{0in}
\item  a  much  improved  mode of  resource  debugging, 
\item  a powerful treatment of multilinguality in linguistic
resources, 
\item and strong
support  for   distributed large-scale  grammar  development.  
\end{itemize}
We  will  briefly note   these features   and  then present some
derivative  functionalities  that also  represent differences  between
analysis   and  generated  oriented GDE's.
For the   functional  approaches, our concrete
descriptions  will be  based on {\sc kpml}:   {\sc fuf} is not
explicitly multilingual and has  as  yet few visualization  tools  for
resource  development  (limited, for example, to  basic  graphs of the
type lattice).  {\sc Kpml} is more similar in its stage of development
to, for example,  {\sc alep},  in that  it  offers a range of  visualisation
techniques for both the static resources and  their dynamic use during
generation, as well   as support  methods for resource   construction,
including    versioning,  resource  merging, and   distinct   kinds of
modularity   for distributed development.    {\sc Fuf} is still mostly
concerned with  the  underlying engine  for   generation
and represents a  programming  environment analogous to
{\sc cuf} or {\sc tdl}.

\subsubsection*{Beyond interactive tracing}

Experiences  with debugging and  maintaining large generation grammars
lead to the conclusion  that `tracing' or `stepping'  during execution
of the resources is usually not a useful way to  proceed. This was the
favored (i.e., only) mode of interaction with, for example, the Penman
system in the 80s.  This has been refined subsequently, both in Penman
and in {\sc kpml} and {\sc fuf}, so that particular tracing can occur,
interactively   if  required, only   when  selected linguistic objects
(e.g.,  particular disjunctions, particular  types of `knowledge base'
access, etc.)  are touched during generation or when particular events
in the generation process occurred.  However, although always necessary
as a  last resort and  for novices, this mode of  debugging has now in
{\sc kpml}   given  way completely  to  `result  focusing'.   Here the
generated  result (which can  be partial in  cases where the resources
fail to produce a  final generated string)  serves as a point of entry
to all decisions taken during generation. This can also be mediated by
the syntactic  structure   generated.    

This is  an effective means  of locating resource problems since, with
the very   conservative `formalism' supported   (see above), there are
only two  possible   sources of generation  errors:  first,  when  the
linguistic resources defined cover the  desired generation result  but
an incorrect grammatical    feature is  selected  (due to    incorrect
semantic  mappings,  or to wrongly constrained  grammatical selections
elsewhere); and second, when the linguistic resources do not cover the
desired  result.  This  means  that the  debugging  task  {\em always}
consists   of  locating where in the   feature  selections made during
generation---i.e.,  in the  selection  expressions  for  the  relevant
grammatical units---an inappropriate selection occurred.

The selection expression list  is accessed from  the user interface by
clicking on any constituent, either from the generated string directly
or from  a graphical representation  of  the syntactic structure.  The
list  itself can  be viewed  in   three ways: (i)   as  a simple  list
functioning as a menu, (ii) as  a graphical representation of the type
lattice (always a  selected subregion of the lattice  as a whole) with
the selected features highlighted,  and (iii) as a  animated graphical
trace of the  `traversal' of the  type lattice  during generation.  In
addition,  all  the  structural  details  of  a generated  string  are
controlled  by  syntactic  constraints  that have  single  determinate
positions  in the type   lattice.  It is  therefore  also  possible to
directly interrogate the    generated string to ask where   particular
structural  features of  the  string were  introduced.  This is a more
focused way of  selecting particular points in the  type  lattice as a
whole for inspection.

Figure~\ref{constraints-to-systems} shows  a  screenshot during   this
latter kind of user activity. The user is attempting  to find out what
where the   lexical constraints responsible for  the  selection of the
noun  ``TIME'' in  the phrase ``At   the same  TIME''  were activated.
Selecting to  see  the   lexical class  constraints  imposed   on this
constituent   ({\tt THING\#3}  in   the structure  top-right) gives  a
listing of  applied  constraints (lower-right).   This indicates which
lexical  constraints    were  applicable   (e.g.,   {\tt  NOUN},  {\tt
COMMON-NOUN},   etc.)   and  {\em   where  in the   type lattice these
constraints  were introduced} (e.g.,  at   the disjunction named  {\tt
HEAD-SUBSTITUTION}, etc.). Clicking  on the disjunction name brings up
a  graphical view of  the  disjunction with the  associated structural
constraints (upper-left). The  feature selected from a  disjunction is
highlighted   in  a  different  color    (or   shade of  grey:    {\tt
lexical-thing}).  The `paradigmatic context' of the disjunction (i.e.,
where in the type lattice it is situated) is  given to the left of the
disjunction proper: this is a boolean  expression over types presented
in standard systemic notation.

Several directions are  then  open to  the user.  The  user can either
follow  the decisions  made  in the  type lattice  to  the left  (less
specific) or to the right (more specific): navigating in either case a
selected subgraph of  the type lattice.   Alternatively,  the user can
inspect the semantic    decisions that   were responsible for      the
particular  selection of grammatical   feature in a disjunction.  This
`upward' move is  also supported graphically. The particular decisions
made and their   paths through  semantic  choice experts  (`choosers')
associated with  each (grammatical) disjunction are shown highlighted.
Since  all objects     presented to the     user are  mouse-sensitive,
navigation and inspection proceeds by direct manipulation. All objects
presented can be edited (either in situ or within automatically linked
Emacs buffers).   Any such changes  are accumulated to define  a patch
version of the loaded resources;   the user can subsequently create  a
distinct  patch for the resources, or  elect to accept  the patches in
the  resource   set.  Generation  itself   is fast  (due to  a  simple
algorithm: see above), and  so   creating a  new `result string'  for
further debugging in the face of changes made is the quickest and most
convenient way to conduct further tests. This  eliminates any need for
backtracking at the user development level. It is possible to examine
contrastively the use of resources across distinct generation cycles.

One useful way of viewing this kind of activity is by contrast to the
state of affairs  when debugging programs.   {\sc Kpml} maintains  the
linguistic  structure as  an   explicit   record  of  the  process  of
generation. All of the decisions  that were made during generation are
accessible via the traces they left in the  generated structure.  Such
information is  typically  not available   when  debugging a  computer
program  since when the  execution stack has been unwound intermediate
results  have  been  lost.   If   certain intermediate  results   must
consequently  be re-examined, it is necessary  to introduce tracing at
appropriate points---a procedure that can now usually be avoided resulting in
significantly faster cycles of debugging/testing.

\subsubsection*{Multilingual representations}

\begin{figure}
\rule{\columnwidth}{0.2mm}
\centerline{\hbox{\psfig{figure=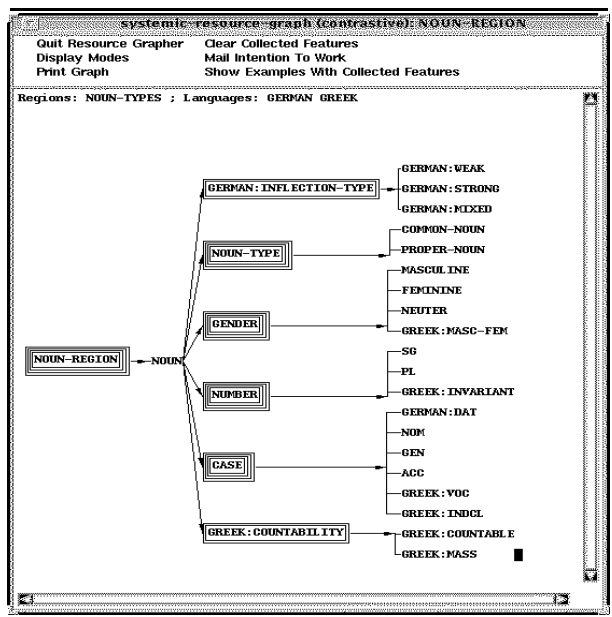,height=80mm,width=80mm}}}
\caption{Views on a multilingual resource}
\label{ml-gr}
\rule{\columnwidth}{0.2mm}
\end{figure}

The use of  multilingual system  networks has  been motivated by,  for
example, Bateman, Matthiessen, Nanri and Zeng~\cite{Bateman-etal91-ijcai}.   {\sc   Kpml}    provides
support for  such resources, including  contrastive graphical displays
of the  type  lattices for distinct languages.    In  addition, it  is
possible to merge  automatically monolingual or  multilingual resource
definitions  and  to separate them out   again as required.  Importing
segments of  a type lattice for  one language to  form a segment for a
distinct language  is also supported.  This  has shown that it  is not
necessary  to maintain a simple  division between, for example, `core'
grammar and variations.  Indeed,   such a division is  wasteful  since
language pairs differ in the  areas they share.   The support for this
multilinguality is organized  entirely  around  the paradigmatic  type
lattice.  The    support   tools   provided  for    manipulating  such
language-conditionalized    lattices   in  {\sc    kpml}    appear  to
significantly reduce the development time for generation resources for
new languages. A black-and-white representation  of a contrastive view
based on   the   Eagles  morphology  recommendations    is  shown   in
Figure~\ref{ml-gr}.  The graph emphasizes areas    held in common  and
explicitly labels  parts of the lattice  that are  restricted in their
language applicability.

The   possibilities  supported    for  working multilingually   (e.g.,
inheritance,   merging  resources)  rely entirely     on the  relative
multilingual applicability  of  the paradigmatic  organization of  the
grammar.  It  appears    a  fact of  multilingual    description  that
paradigmatic  functional  organizations are    more  likely to    show
substantial similarities across  languages  than  are  the syntagmatic
structural  descriptions.  In  an   overview of  resource  definitions
across  6  languages, it was found  that  the multilingual description
only contains 32\% of the number of objects that would  be need if the
6    grammars were represented   separately.   Significant  degrees of
overlap have also been reported whenever a description of one language
has       been         attempted   on        the         basis      of
another (cf.,
e.g.,~\cite{Alshawi-etal92-scle,Rayner-etal96}).    The   paradigmatic  basis  simply   extends  the   range of
similarities that can be represented and provides the formal basis for
providing computational tools that support  the user when constructing
language descriptions `contrastively'.

\subsubsection*{Distributed large-scale grammar development}

The paradigmatic organization of a large-scale grammar shows a further
formal property that is utilized throughout the  {\sc kpml} GDE. Early
work  on systemic  descriptions of  English   noted that emergence  of
`functional   regions': i.e., areas   of the grammar  overall that are
concerned with particular areas of meaning. As Halliday notes:
\begin{quote}
``These [functional] components are reflected in the lexicogrammatical
system in the  form of discrete   networks of options. Each \ldots  is
characterized by strong internal but   weak external constraints:  for
example, any choice made in  transitivity [clause complementation] has
a significant effect on other choices within the transitivity systems,
but  has  very little effect on  choices  within  the mood [speech act
types]         or        theme        [information        structuring]
systems.''~\cite[p113]{Halliday78}.
\end{quote}
This organization was first used computationally in the development of
the Nigel   grammar of English  within the  Penman project.  Nigel was
probably the  first  large-scale computational  grammar  whose precise
authorship is  difficult   to  ascertain  because   of the  number  of
different linguists who have contributed  to it at different times and
locations. 

\begin{figure*}
\rule{\textwidth}{0.2mm}
\centerline{\hbox{\psfig{figure=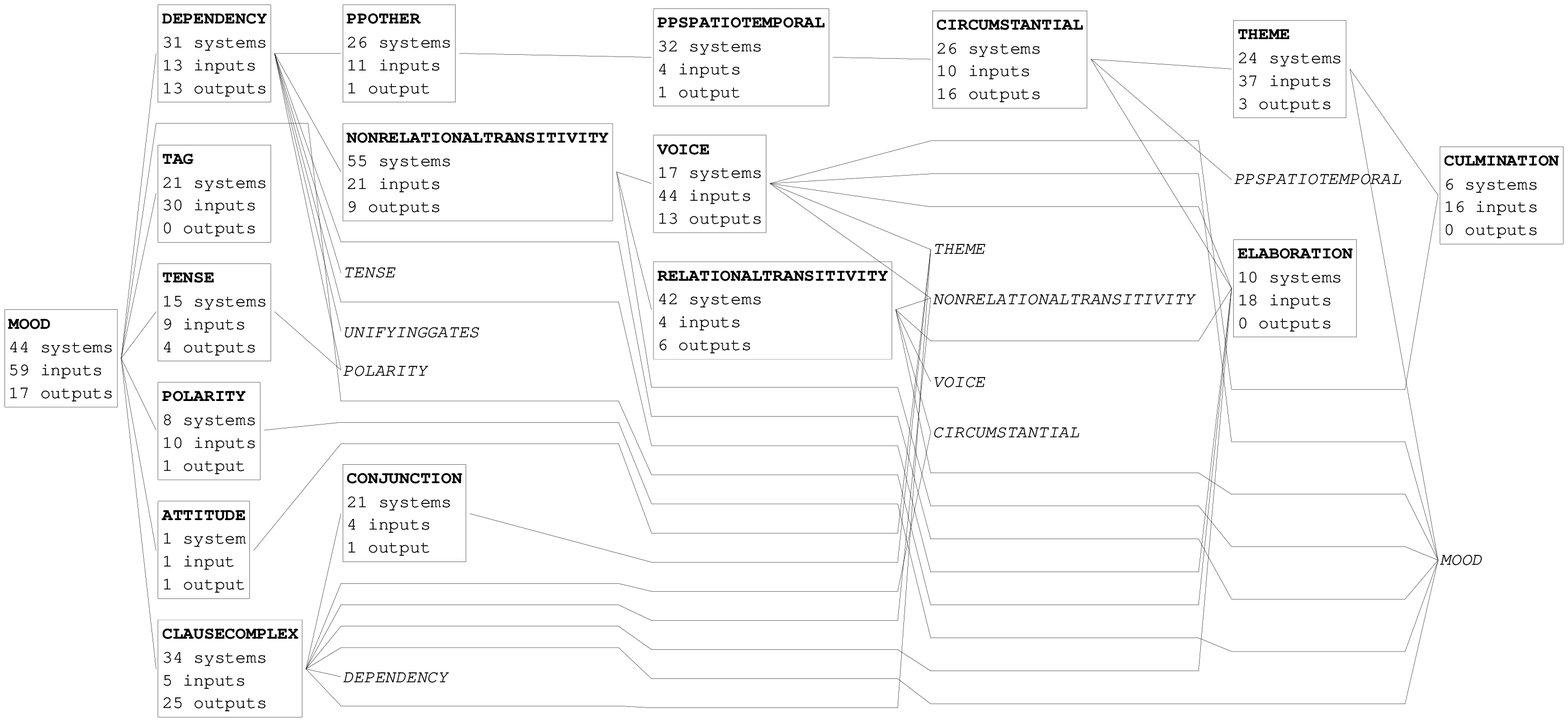,height=80mm,width=130mm}}}
\caption{Functional region connectivity for English (extract)}
\label{region-connectivity}
\rule{\textwidth}{0.2mm}
\end{figure*}

The  basis  for this   successful    example of  distributed   grammar
development is  the organization  of  the overall  type lattice of the
grammar into modular functional regions. This has now  been taken as a
strong design principle within {\sc kpml} where all user access to the
large type lattices making up    a grammar is made through    specific
functional regions: for example, asking  to graph the lattice will  by
default only present information  within a single region (with special
pointers out of the region to indicate broader connectivity).  This is
the paradigmatic equivalent   of maintaining a  structural grammar  in
modules related by  particular syntactic forms.  However, whereas  the
latter information is not strongly organizing for work on a generation
grammar,  the former  is:  work  on  a  generation resource  typically
proceeds by expanding  a selected area of expressive potential---i.e.,
the ability of the grammar to express some  particular set of semantic
requirements. This   can include a range of   grammatical forms and is
best modularized along the paradigmatic    dimension rather than   the
syntagmatic.

The relative strength   of intra-region  connections  in contrast   to
extra-region connections  has provided a  solid basis  for distributed
grammar  development.   Developers  typically announce  that  they are
interested in the expressive potential of some functional region. This
both calls  for others  interested  in the same functional   region to
exchange results cooperatively and warns  generally that a  functional
region  may be subject to imminent  change. When a  revised version of
the  region   is   available    it replaces  the     previous  version
used.  Integration of the new  region  is facilitated  by  a range  of
visualization  tools  and   connectivity checks:  the   final test  of
acceptability  is that all test suites  (see next subsection) generate
the  same results as  with the previous  region version and that a new
test suite  is provided  that  demonstrates  the increased  or revised
functionality of the new region.

Regions  are  defined   across  languages:  the  current  multilingual
resources released with {\sc kpml} include around 60 regions. A partial region
connectivity   graph  for  the       English grammar is    shown    in
Figure~\ref{region-connectivity}. This  graph also serves  as a `menu'
for accessing further  graphical views of the type  lattice as well as
selections    from  test suites  illustrating    use  of the resources
contained within  a  region.   Dependencies between  regions are  thus
clearly indicated.

\subsubsection*{Integrated test suites}

Sets  of linguistic resources for   generation are typically  provided
with test  suites: such  test suites  consist minimally  of a semantic
specification and the string  that  should result when generating.  In
{\sc kpml}   these are indexed  according to  the grammatical features
that are  selected during their  generation. This permits  examples of
the use and consequences  of any feature  from the type lattice to  be
presented during debugging. This is one particularly effective way not
only of checking the status of resources but  also for documenting the
resources. The complete generation history of examples can be examined
in exactly the  same way as   newly generated strings.  An interesting
line of development underway is to investigate correspondences between
the paradigmatic features   describing features in a {\sc kpml}-example  set
and those features used in the TSNLP initiative.

\section{Discussion}

The basic premises of a generation-oriented GDE such as {\sc kpml} differ in
certain respects to  those those of an  analysis-oriented  GDE such as
{\sc alep}.   This  also  stretches to  the  style of   interaction with the
system. For example,  interaction with the {\sc Kpml}  GDE is, as with
Smalltalk  and {\sc alep}, {\em object-oriented}   but, in contrast to {\sc alep},
the objects to which a user has access are strongly restricted to just
those  {\em linguistic constructs} that   are relevant for generation.
This separates development environment details from the resources that
are being  developed. This is, of  course, both possible and desirable
because {\sc kpml} is not intended to be tailored for particular types
of resource by the user: the theoretical orientation is fixed.

The benefits    of this approach seem to    far outweigh  the apparent
limitations.  First, the visualizations  provided are exactly tailored
to the details  of the linguistic objects supported  and their use  in
generation.   Thus  resource sets, networks,  systems  (disjunctions),
semantic choice experts,  dynamic traversal of the network,  syntactic
structures,   etc.  all    have    their  own  distinctive   graphical
representations: this establishes a clear modularity in the conception
of  the user  that is easily   obscured when a single more  generation
representation style  (e.g., feature structure  presented in a feature
structure editor)  is  used for a wide   range  of information.   This
clarifies the difference     in information modules    and thus  helps
development.  It is then also possible  to `fold' generation decisions
into the visualizations in a natural way:  thus supporting the `result
focusing'   mode  of development  described  above.    Thus,  whenever
resources are inspected,   their use    during  selected cycles     of
generation  is  also  displayed   by  highlighting or  annotating  the
appropriate objects shown.

This  also  influences the  {\em  kind} of user for  which  the GDE is
appropriate.  The central  areas in   generation are still   primarily
functional and pragmatic  rather than structural  and syntactic. It is
less common that   linguists and developers concerned with  pragmatics
and text   linguistics are  fully   comfortable with  constraint logic
programming.    The  dedicated graphical  presentation  of  linguistic
objects provided  in {\sc  kpml}  appears to  provide a more generally
accessible  tool  for   constructing linguistic descriptions.  Grammar
components have been  written using {\sc  kpml} by computer scientists
without  training  in computational   linguistics,  by functional text
linguists,  by  translators and   technical  writers, as  well  as  by
computational and systemic-functional linguists. 

Finally,      however,   the  well     understood relationship between
systemic-functional style descriptions and, for example, typed feature
representations   provides  a bridge    from  the  less   formal, more
functional style of  description  back to the kind  of representations
found in  NLA-oriented GDE's. It  is therefore to   be expected that a
broader  range of  linguistic   input and  development   work  will be
encouraged to feed into large-scale resource development than would be
possible if the kind of development were limited to that practised for
purposes of analysis.


{\small
\begin{thebibliography}{}

\bibitem[\protect\citename{Alshawi \bgroup et al.\egroup
  }1992]{Alshawi-etal92-scle}
Hiyan Alshawi, David Carter, {Bj\"orn} {Gamb\"ack}, and Manny Rayner.
\newblock 1992.
\newblock {Swedish-English QLF translation}.
\newblock In Hiyan Alshawi, editor, {\em The Core Language Engine}, pages 277
  -- 319. MIT Press.

\bibitem[\protect\citename{Appelt}1985]{Appelt85}
Douglas~E. Appelt.
\newblock 1985.
\newblock {\em Planning Natural Language Utterances}.
\newblock Cambridge University Press, Cambridge, England.

\bibitem[\protect\citename{Bateman \bgroup et al.\egroup
  }1991]{Bateman-etal91-ijcai}
John~A. Bateman, Christian~M.I.M. Matthiessen, Keizo Nanri, and Licheng Zeng.
\newblock 1991.
\newblock The re-use of linguistic resources across languages in multilingual
  generation components.
\newblock In {\em Proceedings of the 1991 International Joint Conference on
  Artificial Intelligence, Sydney, Australia}, volume~2, pages 966 -- 971.
  Morgan Kaufmann Publishers.

\bibitem[\protect\citename{Bateman}1997]{KPML1}
John~A. Bateman, 1997.
\newblock {\em {KPML Development Environment: multilingual linguistic resource
  development and sentence generation}}.
\newblock {German National Center for Information Technology (GMD), Institute
  for integrated publication and information systems (IPSI)}, Darmstadt,
  Germany, January.
\newblock (Release 1.1).

\bibitem[\protect\citename{Bateman}to appear]{Bateman-ADG}
John~A. Bateman.
\newblock to appear.
\newblock Automatic discourse generation.
\newblock In Allen Kent, editor, {\em Encyclopedia of Library and Information
  Science}. Marcel Dekker, Inc., New York.

\bibitem[\protect\citename{{EAGLES}}1996]{Eagles-fwg96}
{EAGLES}.
\newblock 1996.
\newblock Formalisms working group final report.
\newblock Expert advisory group on language engineering standards document,
  September.

\bibitem[\protect\citename{Elhadad}1990]{Elhadad90-types}
Michael Elhadad.
\newblock 1990.
\newblock Types in functional unification grammars.
\newblock In {\em Proceedings of the 28th. Annual Meeting of the Association
  for Computational Linguistics}, pages 157 --164. Association for
  Computational Linguistics.

\bibitem[\protect\citename{Halliday}1978]{Halliday78}
Michael~A.K. Halliday.
\newblock 1978.
\newblock {\em Language as social semiotic}.
\newblock Edward Arnold, London.

\bibitem[\protect\citename{Mann and Matthiessen}1985]{MannMatthiessen85-benson}
William~C. Mann and Christian~M.I.M. Matthiessen.
\newblock 1985.
\newblock Demonstration of the {N}igel text generation computer program.
\newblock In James~D. Benson and William~S. Greaves, editors, {\em Systemic
  Perspectives on Discourse, Volume 1}, pages 50--83. Ablex, Norwood, New
  Jersey.

\bibitem[\protect\citename{McDonald}1980]{McDonald80}
David~D. McDonald.
\newblock 1980.
\newblock {\em Natural Language Production as a Process of Decision Making
  under Constraint}.
\newblock {Ph.D.} thesis, MIT, Cambridge, Mass.

\bibitem[\protect\citename{McKeown}1985]{McKeown85-book}
Kathleen~R McKeown.
\newblock 1985.
\newblock {\em Text Generation: Using Discourse Strategies and Focus
  Constraints to Generate Natural Language Text}.
\newblock Cambridge University Press, Cambridge, England.

\bibitem[\protect\citename{Mel'\v{c}uk and
  \v{Z}holkovskij}1970]{MelcukZholkovskij70}
A.~Mel'\v{c}uk, Igor and A.K. \v{Z}holkovskij.
\newblock 1970.
\newblock Towards a functioning ``meaning-text'' model of language.
\newblock {\em Linguistics}, 57:10--47.

\bibitem[\protect\citename{Pulman}1991]{ET-6-finalreport}
Stephen~G. Pulman, editor.
\newblock 1991.
\newblock {\em {\sc Eurotra ET6/1}: rule formalism and virtual machine design
  study -- final report}.
\newblock Commission of the European Communities, Luxembourg.
\newblock Contributors: H. Alshawi, D.J. Arnold, R. Backofen, D.M. Carter, J.
  Lindop, K. Netter, S.G. Pulman, J. Tsujii and H. Uszkoreit.

\bibitem[\protect\citename{Rayner \bgroup et al.\egroup }1996]{Rayner-etal96}
M.~Rayner, D.~Carter, and P.~Bouillon.
\newblock 1996.
\newblock Adapting the core language engine to french and spanish.
\newblock In {\em Proceedings of NLP-IA-96}, Moncton, new Brunswick, May.

\bibitem[\protect\citename{{Sch\"utz}}1996]{Schuetz96-amta}
{J\"org} {Sch\"utz}.
\newblock 1996.
\newblock {European Research and Development in Machine Translation}.
\newblock {\em {MT News International}}, 15:8--11, October.
\newblock (Newsletter of the International Association for Machine
  Translation).

\bibitem[\protect\citename{Simpkins \bgroup et al.\egroup
  }1993]{Simpkins-etal93}
N.K. Simpkins, G.~Cruickshank, and {P.E International}.
\newblock 1993.
\newblock {ALEP-0 Virtual Machine extensions}.
\newblock Technical report, CEC.

\bibitem[\protect\citename{Zock and Adorni}1996]{ZockAdorni96-intro}
Michael Zock and Giovanni Adorni.
\newblock 1996.
\newblock Introduction.
\newblock In Giovanni Adorni and Michael Zock, editors, {\em Trends in natural
  language generation: an artificial intelligence perspective}, number 1036 in
  Lecture Notes in Artificial Intelligence, pages 1--16. Springer-Verlag.

\end{thebibliography}}
\end{document}